\documentclass[conference]{IEEEtran}
\IEEEoverridecommandlockouts
\usepackage{cite}
\usepackage{amsmath,amssymb,amsfonts}
\usepackage{algorithmic}
\usepackage{graphicx}
\usepackage{textcomp}
\usepackage{xcolor}
\usepackage{subcaption}
\usepackage{multirow}
\usepackage{booktabs}

\def\BibTeX{{\rm B\kern-.05em{\sc i\kern-.025em b}\kern-.08em
    T\kern-.1667em\lower.7ex\hbox{E}\kern-.125emX}}
\begin{document}

\title{Efficient Multi-scale Masked Autoencoders \\with Hybrid-Attention Mechanism \\for Breast Lesion Classification}

\makeatletter
\newcommand{\linebreakand}{%
  \end{@IEEEauthorhalign}
  \hfill\mbox{}\par
  \mbox{}\hfill\begin{@IEEEauthorhalign}
}
\makeatother

\author{\IEEEauthorblockN{Hung Q. Vo\IEEEauthorrefmark{2},
Pengyu Yuan\IEEEauthorrefmark{2}, Zheng Yin\IEEEauthorrefmark{3}, Kelvin K. Wong\IEEEauthorrefmark{3},\\ Chika F. Ezeana\IEEEauthorrefmark{3}, Son T. Ly\IEEEauthorrefmark{2}, Hien V. Nguyen\IEEEauthorrefmark{3} and Stephen T.C. Wong\IEEEauthorrefmark{2}
}
\thanks{* Corresponding email: hqvo2@uh.edu}
\thanks{** Hien V. Nguyen and Stephen T.C. Wong are co-senior authors}
\IEEEauthorblockA{\IEEEauthorrefmark{2}Department of Electrical and Computer Engineering,
University of Houston\\
\IEEEauthorrefmark{3}Systems Medicine and Biomedical Engineering,
Houston Methodist}}

\let\oldtwocolumn\twocolumn

\maketitle

\begin{abstract}
Self-supervised learning (SSL) with Vision Transformers (ViT) has shown immense potential in medical image analysis. However, the quadratic complexity ($\mathcal{O}(N^2)$) of standard self-attention poses a severe barrier for high-resolution biomedical tasks, effectively excluding resource-constrained research labs from utilizing state-of-the-art models. To address this computational bottleneck without sacrificing diagnostic accuracy, we propose \textbf{MIRAM}, a Multi-scale Masked Autoencoder that leverages a \textbf{hybrid-attention mechanism}. 

Our architecture uniquely decouples semantic learning from detail reconstruction using a dual-decoder design: a standard transformer decoder captures global semantics at low resolution, while a linear-complexity decoder (comparing Linformer, Performer, and Nyströmformer) handles the computationally expensive high-resolution reconstruction. This reduces the complexity of the upscaling stage from quadratic to linear ($\mathcal{O}(N)$), enabling high-fidelity training on consumer-grade GPUs. We validate our approach on the CBIS-DDSM mammography dataset. Remarkably, our \textbf{Nyströmformer-based variant} achieves a classification accuracy of \textbf{61.0\%}, outperforming both standard MAE (58.9\%) and MoCo-v3 (60.2\%) while requiring significantly less memory. These results demonstrate that hybrid-attention architectures can democratize high-resolution medical AI, making powerful SSL accessible to researchers with limited hardware resources.
\end{abstract}

\begin{IEEEkeywords}
Self-supervised Learning, Hybrid-Attention Mechanism, Masked Autoencoder, Breast Cancer, Mammogram.
\end{IEEEkeywords}

\section{Introduction}

The scarcity of expert-annotated data remains a primary bottleneck in the deployment of deep learning for medical diagnostics. While obtaining pixel-perfect annotations from radiologists is expensive and time-consuming, hospitals generate vast quantities of unannotated medical imaging data daily. Conventional supervised learning leaves this resource untapped. Self-Supervised Learning (SSL) has emerged as a promising paradigm to bridge this gap, enabling models to learn robust feature representations from unlabeled data by exploiting inherent structures or cross-modal relationships \cite{azizi2021big, li2021dual}.

In the domain of breast cancer diagnosis via mammograms (MGs), SSL has shown initial promise. Recent works have utilized contrastive learning \cite{chen2021empirical} or self-distillation \cite{caron2021emerging} to classify whole mammograms or ultrasound images \cite{perek2021self}. However, these dominant SSL paradigms rely on data augmentation strategies that encourage invariance to scaling and spatial shifting. While effective for global classification, this invariance can be detrimental for fine-grained pathology tasks. In mammography, the distinction between a benign and malignant lesion often lies in minute, high-frequency details—such as microcalcifications or subtle margin irregularities—that contrastive methods may discard as "noise."

To capture these fine-grained details, generative approaches like the Masked Autoencoder (MAE) \cite{he2021masked} have gained traction. By masking random patches of an image and forcing the model to reconstruct the missing pixels, MAE compels the network to learn a deep understanding of spatial semantics. However, applying standard MAE to high-resolution medical images presents a severe computational challenge. Reconstructing high-fidelity details requires high-resolution inputs and heavy decoders. Since the attention mechanism in standard Transformers scales quadratically ($\mathcal{O}(N^2)$) with sequence length, training such models on high-resolution medical scans is often computationally prohibitive for research labs with limited GPU resources.

To address these challenges, we introduce \textbf{MIRAM} (Masked Image Reconstruction Across Multiple scales), a resource-efficient SSL framework designed for lesion-level analysis. Unlike standard MAE, which operates at a single scale, MIRAM utilizes a novel dual-decoder architecture to perform multi-scale inpainting. A standard transformer decoder reconstructs low-resolution global semantics, while a second, efficient decoder handles high-resolution details. Crucially, we investigate the integration of linear-complexity attention mechanisms—specifically Linformer, Performer, and Nyströmformer—into the high-resolution decoder. This hybrid design allows us to model intricate spatial details without the memory explosion associated with full self-attention, democratizing high-resolution pre-training for resource-constrained environments.

Our contributions are as follows:
\begin{itemize}
    \item We propose a novel pretext task, Multi-Scale Masked Reconstruction, which forces the model to simultaneously learn global semantic structure and fine-grained local texture, crucial for lesion risk analysis.
    \item We introduce a hybrid-attention architecture that reduces the complexity of the high-resolution decoder from $\mathcal{O}(N^2)$ to $\mathcal{O}(N)$, effectively enabling high-resolution training on consumer-grade GPUs.
    \item We demonstrate that our efficient Nyströmformer variant achieves superior performance on the CBIS-DDSM dataset, outperforming standard MAE and supervised baselines in pathology classification (61.0\% vs 58.9\% accuracy) while significantly reducing computational costs.
\end{itemize}

\begin{figure}[!]  
    \centering  
    \begin{subfigure}{0.45\textwidth}  
        \centering  
        \subcaption[]{Mass Lesion}  
        \begin{subfigure}{0.18\textwidth}  
            \includegraphics[width=\textwidth]{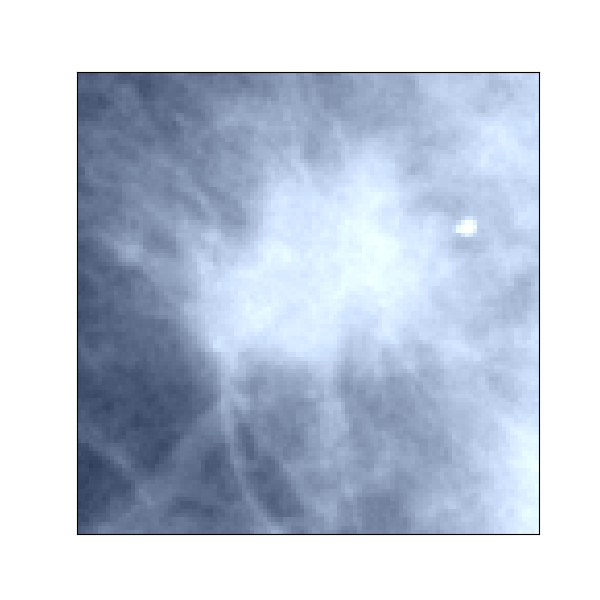}  
            \captionsetup{justification=centering, font=scriptsize}  
        \end{subfigure}  
        \hspace{-0.3cm}  
        \begin{subfigure}{0.25\textwidth}  
            \includegraphics[width=\textwidth]{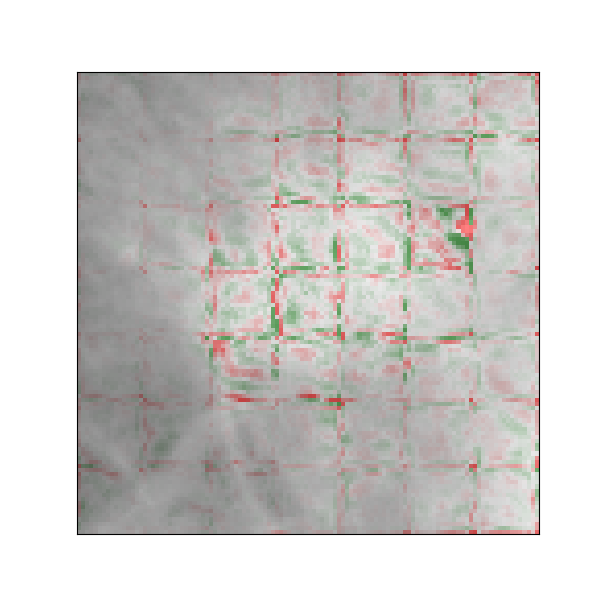}  
            \captionsetup{justification=centering, font=scriptsize}  
        \end{subfigure}  
        \hspace{-0.5cm}  
        \begin{subfigure}{0.25\textwidth}  
            \includegraphics[width=\textwidth]{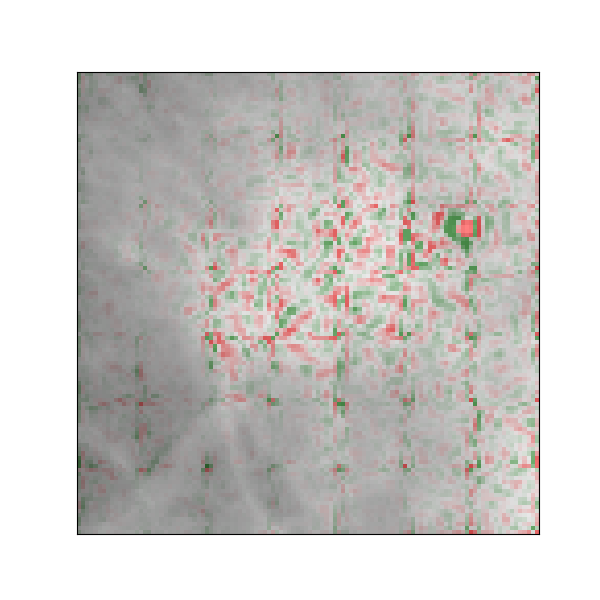}  
            \captionsetup{justification=centering, font=scriptsize}  
        \end{subfigure}  
        \begin{subfigure}{0.14\textwidth}  
            \includegraphics[width=\textwidth]{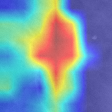}  
            \captionsetup{justification=centering, font=scriptsize}  
        \end{subfigure}  
        \begin{subfigure}{0.14\textwidth}  
            \includegraphics[width=\textwidth]{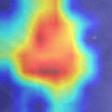}  
            \captionsetup{justification=centering, font=scriptsize}  
        \end{subfigure}        

        \begin{subfigure}{0.18\textwidth}  
            \includegraphics[width=\textwidth]{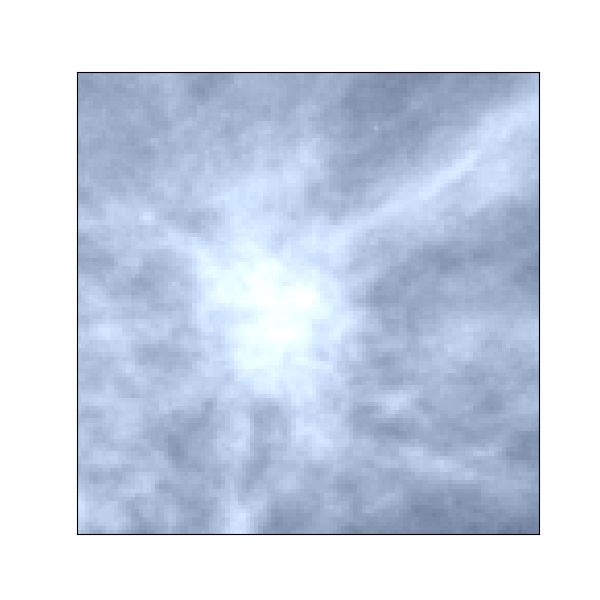}  
            \captionsetup{justification=centering, font=scriptsize}  
        \end{subfigure}  
        \hspace{-0.3cm}  
        \begin{subfigure}{0.25\textwidth}  
            \includegraphics[width=\textwidth]{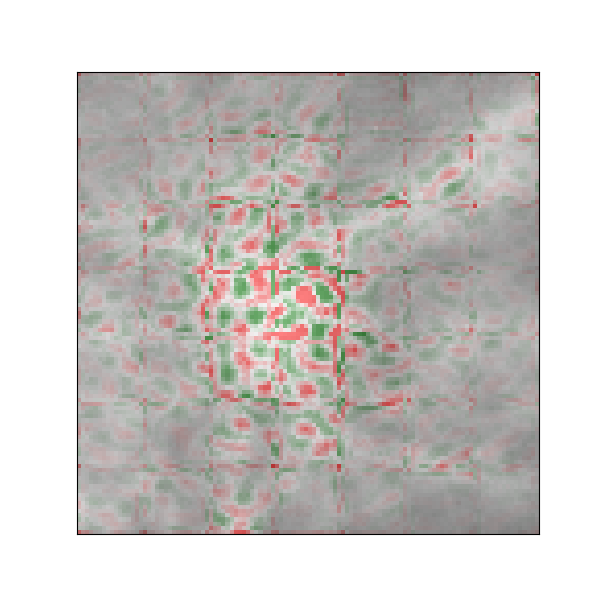}  
            \captionsetup{justification=centering, font=scriptsize}  
        \end{subfigure}  
        \hspace{-0.5cm}  
        \begin{subfigure}{0.25\textwidth}  
            \includegraphics[width=\textwidth]{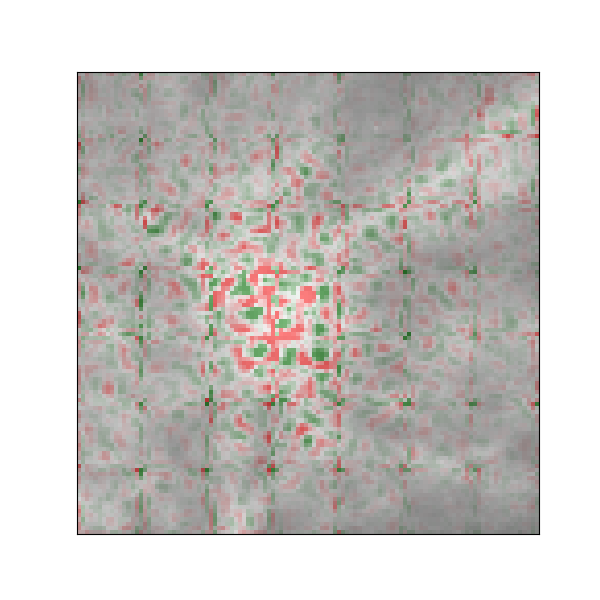}  
            \captionsetup{justification=centering, font=scriptsize}  
        \end{subfigure}  
        \begin{subfigure}{0.14\textwidth}  
            \includegraphics[width=\textwidth]{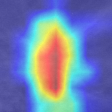}  
            \captionsetup{justification=centering, font=scriptsize}  
        \end{subfigure}  
        \begin{subfigure}{0.14\textwidth}  
            \includegraphics[width=\textwidth]{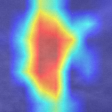}  
            \captionsetup{justification=centering, font=scriptsize}  
        \end{subfigure} 

        \begin{subfigure}{0.18\textwidth}  
            \vspace{0.3cm}  
            \includegraphics[width=\textwidth]{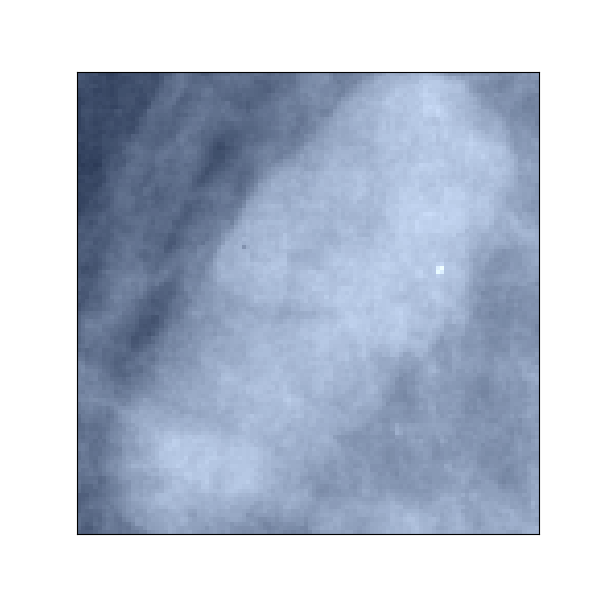}  
            \captionsetup{justification=centering, font=scriptsize}  
            \caption*{Input Image}  
        \end{subfigure}  
        \hspace{-0.3cm}  
        \begin{subfigure}{0.25\textwidth}  
            \vspace{0.3cm}  
            \includegraphics[width=\textwidth]{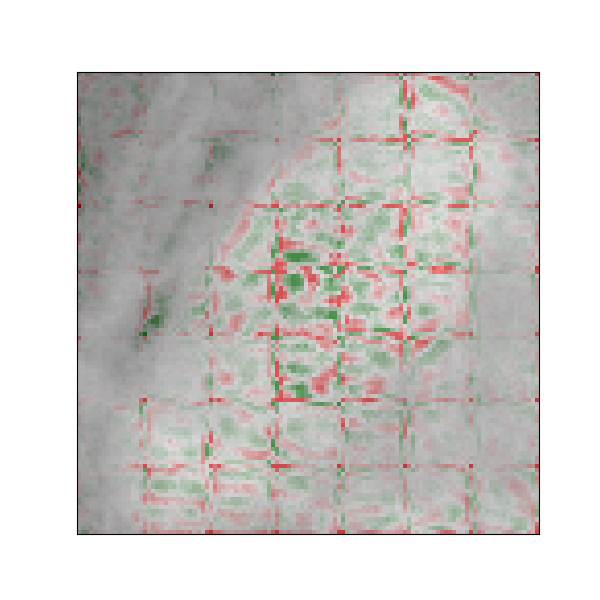}  
            \captionsetup{justification=centering, font=scriptsize}  
            \caption*{DeepLift\\(MAE-Baseline)}  
        \end{subfigure}  
        \hspace{-0.5cm}  
        \begin{subfigure}{0.25\textwidth}  
            \vspace{0.3cm}  
            \includegraphics[width=\textwidth]{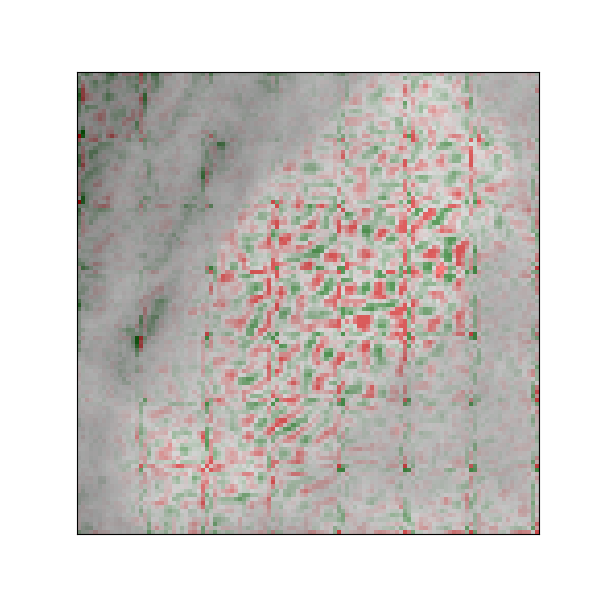}  
            \captionsetup{justification=centering, font=scriptsize}  
            \caption*{DeepLift\\(MIRAM-Ours)}  
        \end{subfigure}  
        \begin{subfigure}{0.14\textwidth}  
            \vspace{0.3cm}  
            \includegraphics[width=\textwidth]{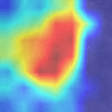}  
            \captionsetup{justification=centering, font=scriptsize}  
            \caption*{ScoreCAM\\(MAE-Baseline)}  
        \end{subfigure}  
        \begin{subfigure}{0.14\textwidth}  
            \vspace{0.3cm}  
            \includegraphics[width=\textwidth]{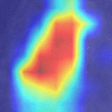}  
            \captionsetup{justification=centering, font=scriptsize}  
            \caption*{ScoreCAM\\(MIRAM-Ours)}  
        \end{subfigure}  
    \end{subfigure}   

    \vspace{0.5cm} 

    \begin{subfigure}{0.45\textwidth}  
        \centering  
        \subcaption[]{Calcification Lesion}  

        \begin{subfigure}{0.18\textwidth}  
            \includegraphics[width=\textwidth]{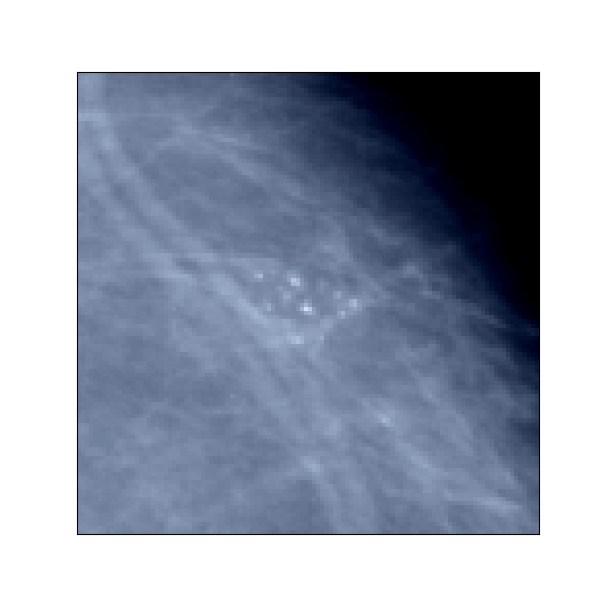}  
            \captionsetup{justification=centering, font=scriptsize}  
        \end{subfigure}  
        \hspace{-0.3cm}  
        \begin{subfigure}{0.25\textwidth}  
            \includegraphics[width=\textwidth]{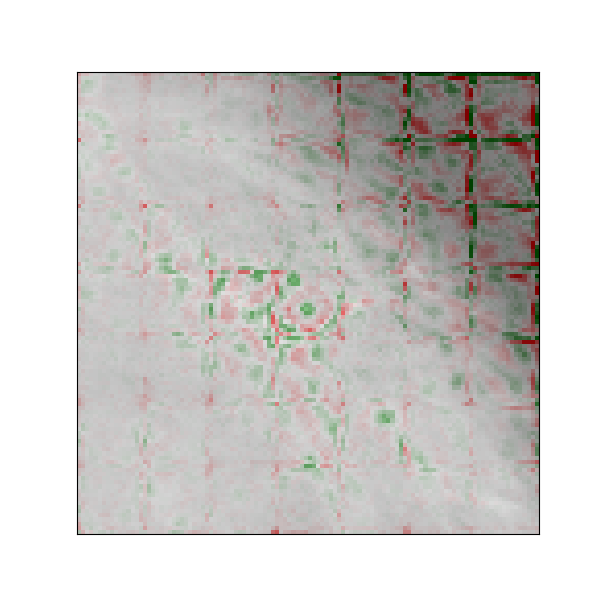}  
            \captionsetup{justification=centering, font=scriptsize}  
        \end{subfigure}  
        \hspace{-0.5cm}  
        \begin{subfigure}{0.25\textwidth}  
            \includegraphics[width=\textwidth]{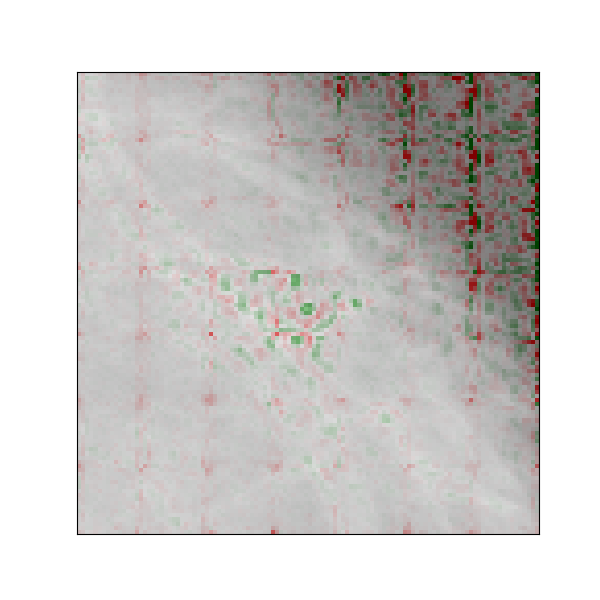}  
            \captionsetup{justification=centering, font=scriptsize}  
        \end{subfigure}  
        \begin{subfigure}{0.14\textwidth}  
            \includegraphics[width=\textwidth]{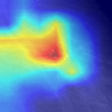}  
            \captionsetup{justification=centering, font=scriptsize}  
        \end{subfigure}  
        \begin{subfigure}{0.14\textwidth}  
            \includegraphics[width=\textwidth]{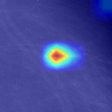}  
            \captionsetup{justification=centering, font=scriptsize}  
        \end{subfigure}

        \begin{subfigure}{0.18\textwidth}  
            \includegraphics[width=\textwidth]{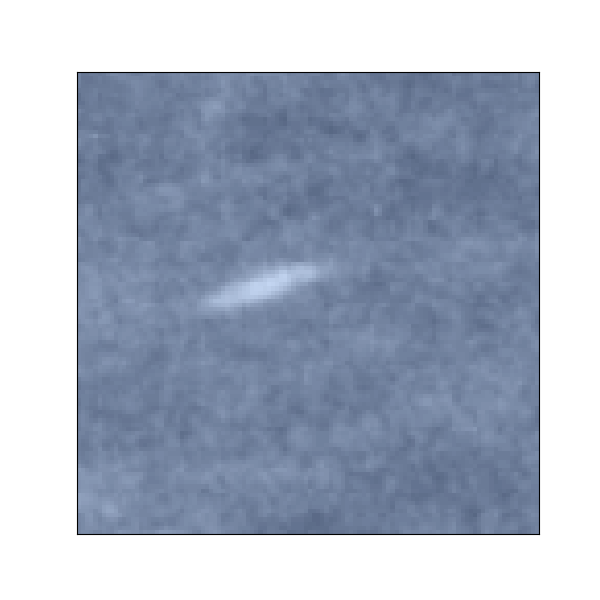}  
            \captionsetup{justification=centering, font=scriptsize}  
        \end{subfigure}  
        \hspace{-0.3cm}  
        \begin{subfigure}{0.25\textwidth}  
            \includegraphics[width=\textwidth]{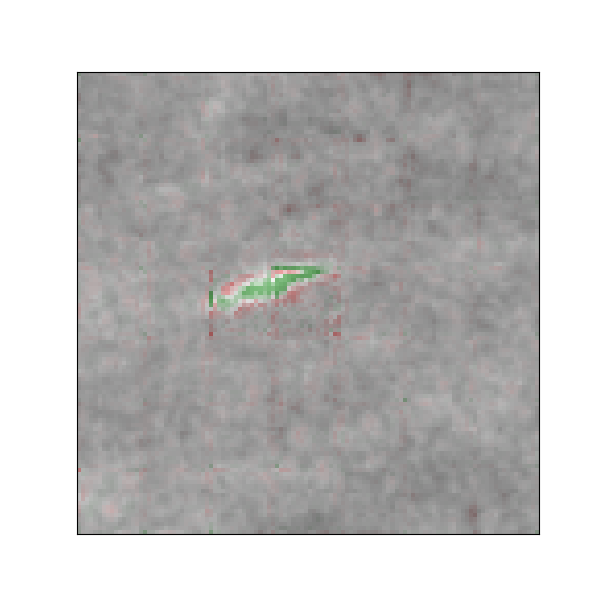}  
            \captionsetup{justification=centering, font=scriptsize}  
        \end{subfigure}  
        \hspace{-0.5cm}  
        \begin{subfigure}{0.25\textwidth}  
            \includegraphics[width=\textwidth]{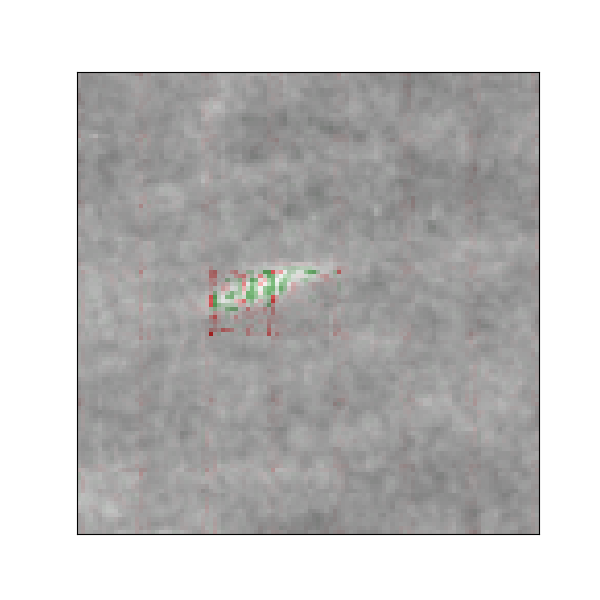}  
            \captionsetup{justification=centering, font=scriptsize}  
        \end{subfigure}  
        \begin{subfigure}{0.14\textwidth}  
            \includegraphics[width=\textwidth]{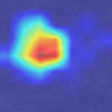}  
            \captionsetup{justification=centering, font=scriptsize}  
        \end{subfigure}  
        \begin{subfigure}{0.14\textwidth}  
            \includegraphics[width=\textwidth]{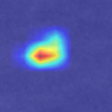}  
            \captionsetup{justification=centering, font=scriptsize}  
        \end{subfigure}

        \begin{subfigure}{0.18\textwidth}  
            \vspace{0.3cm}  
            \includegraphics[width=\textwidth]{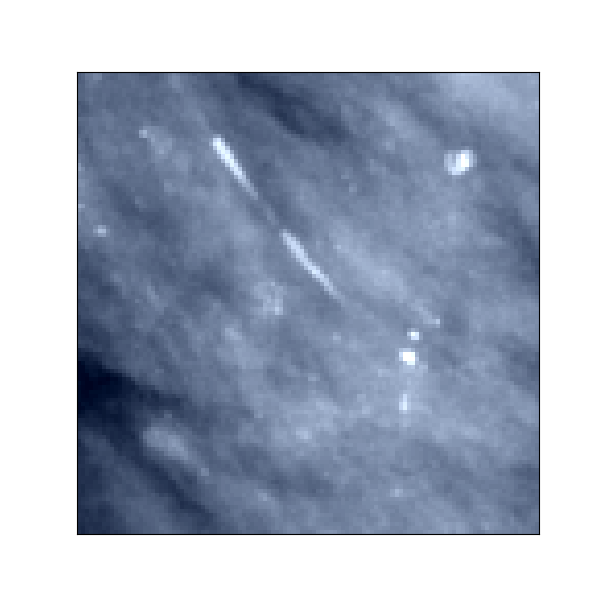}  
            \captionsetup{justification=centering, font=scriptsize}  
            \caption*{Input Image}  
        \end{subfigure}  
        \hspace{-0.3cm}  
        \begin{subfigure}{0.25\textwidth}  
            \vspace{0.3cm}  
            \includegraphics[width=\textwidth]{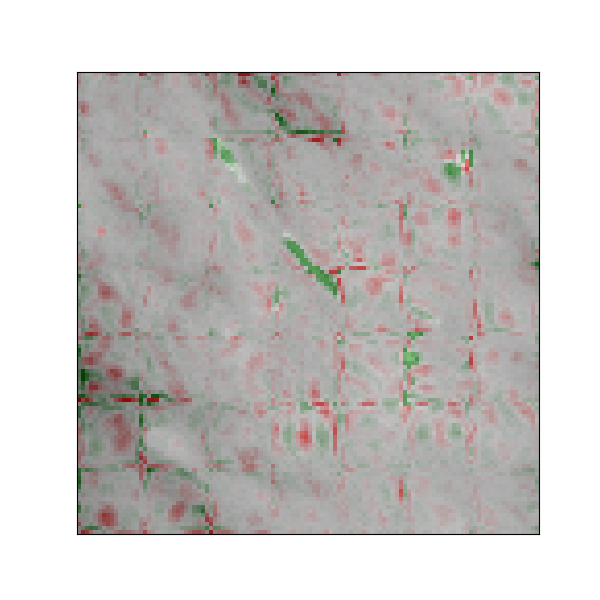}  
            \captionsetup{justification=centering, font=scriptsize}  
            \caption*{DeepLift\\(MAE-Baseline)}  
        \end{subfigure}  
        \hspace{-0.5cm}  
        \begin{subfigure}{0.25\textwidth}  
            \vspace{0.3cm}  
            \includegraphics[width=\textwidth]{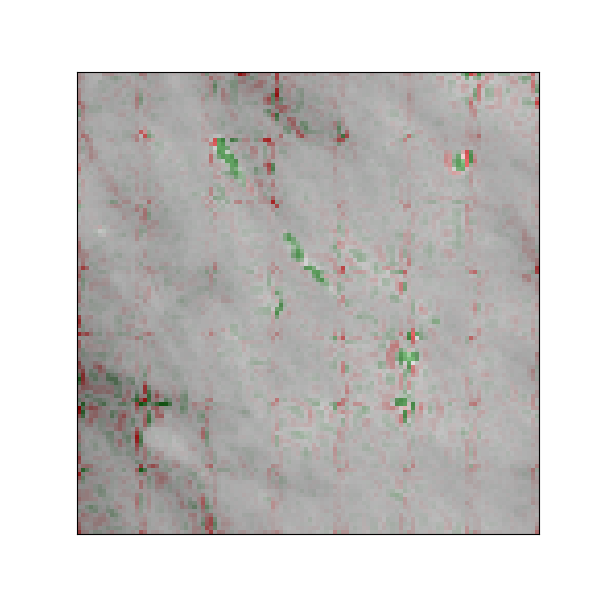}  
            \captionsetup{justification=centering, font=scriptsize}  
            \caption*{DeepLift\\(MIRAM-Ours)}  
        \end{subfigure}  
        \begin{subfigure}{0.14\textwidth}  
            \vspace{0.3cm}  
            \includegraphics[width=\textwidth]{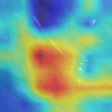}  
            \captionsetup{justification=centering, font=scriptsize}  
            \caption*{ScoreCAM\\(MAE-Baseline)}  
        \end{subfigure}  
        \begin{subfigure}{0.14\textwidth}  
            \vspace{0.3cm}  
            \includegraphics[width=\textwidth]{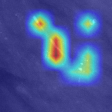}  
            \captionsetup{justification=centering, font=scriptsize}  
            \caption*{ScoreCAM\\(MIRAM-Ours)}  
        \end{subfigure}  
    \end{subfigure}  

    \caption{Feature visualization on the pathology classification task. MIRAM outperforms the original MAE by more accurately targeting the lesion region and capturing subtle fine-grained details.}  
    \vspace{-6mm}  
    \label{fig:vis_feat}  
\end{figure}

\begin{table*}[t]
\centering
\caption{\textbf{Computational complexity analysis of the Dual-Decoder MAE architecture.} \\
Decoder 1 reconstructs the original resolution ($K \times K$) from sequence length $N$. Decoder 2 reconstructs high-resolution ($2K \times 2K$) from sequence length $4N$. $d$ denotes the embedding dimension, and $m$ represents the projection dimension or number of landmarks/random features for linear approximations (where $m \ll N$). Note the quadratic explosion in complexity for the Standard Transformer in Decoder 2.}
\label{tab:complexity_analysis}
\resizebox{0.9\textwidth}{!}{%
\begin{tabular}{@{}llcccc@{}}
\toprule
\textbf{Module} & \textbf{Attention Mechanism} & \textbf{Sequence Length} & \textbf{Time Complexity} & \textbf{Space Complexity} & \textbf{Relative Cost} \\ 
& & & & & \textit{(vs. Decoder 1)} \\ \midrule
\textbf{Decoder 1} & Standard Transformer & $N$ & $\mathcal{O}(N^2 d)$ & $\mathcal{O}(N^2 + Nd)$ & $1\times$ \\ \midrule
\multirow{4}{*}{\textbf{Decoder 2}} & Standard Transformer & $4N$ & $\mathcal{O}(16 N^2 d)$ & $\mathcal{O}(16 N^2 + 4Nd)$ & \textbf{16$\times$ (Quadra. Expl.)} \\ \cmidrule(l){2-6} 
 & Linformer & $4N$ & $\mathcal{O}(4 Nmd)$ & $\mathcal{O}(4 Nm + 4Nd)$ & \textbf{4$\times$} \\
 & Performer & $4N$ & $\mathcal{O}(4 Nmd)$ & $\mathcal{O}(4 Nmd + d^2)$ & \textbf{4$\times$} \\
 & Nyströmformer & $4N$ & $\mathcal{O}(4 Nmd)$ & $\mathcal{O}(4 Nmd + m^2d)$ & \textbf{4$\times$} \\ \bottomrule
\end{tabular}%
}
\end{table*}

\section{Methodology}

\begin{figure*}[ht!]  
    \centering  
    \includegraphics[width=\textwidth]{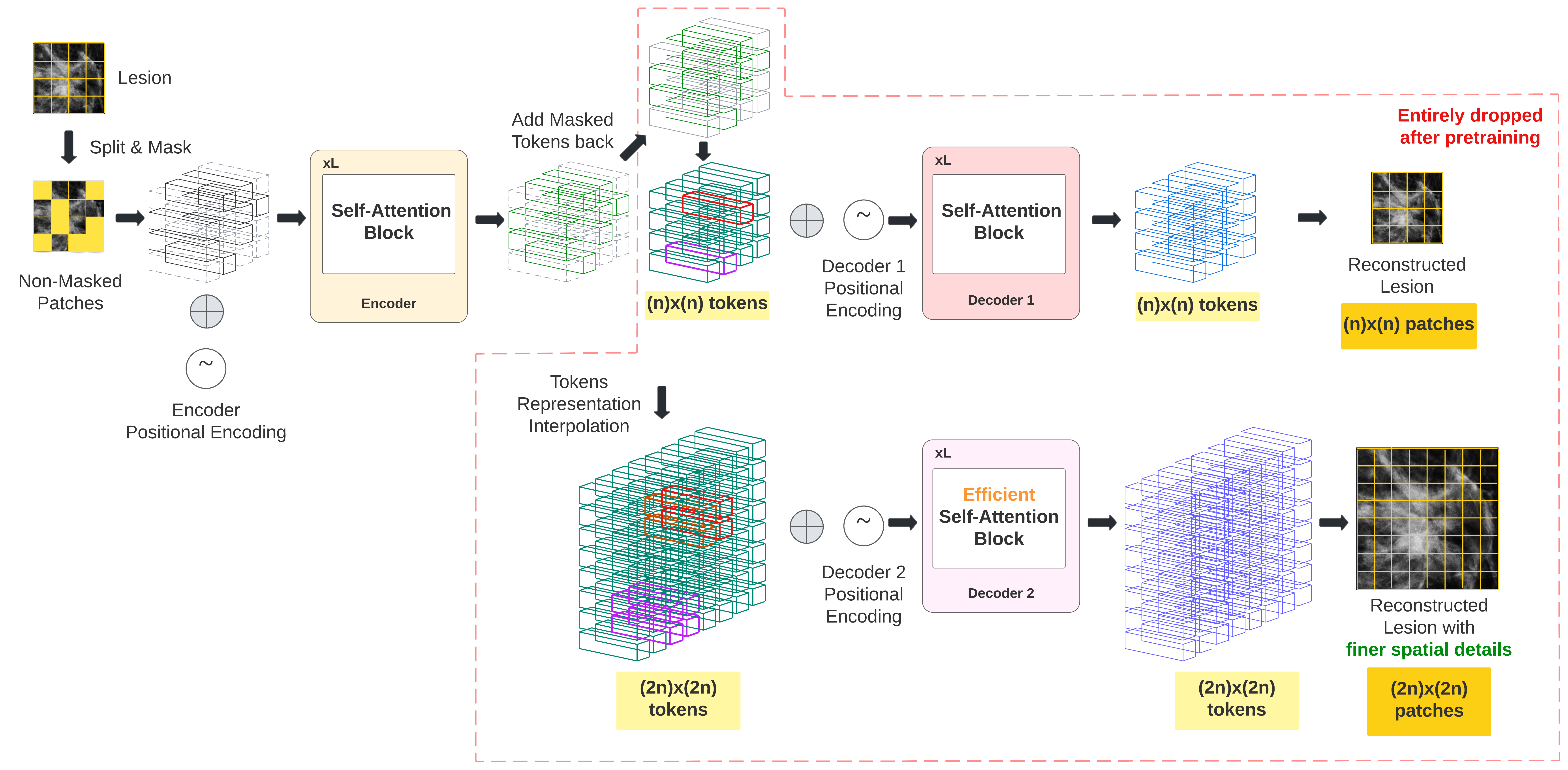}  
    \caption{Overview of the MIRAM SSL Framework. The model utilizes a dual-decoder design to reconstruct the image at two distinct scales: the original input resolution and a $2\times$ spatially upsampled high-resolution output.}
    \vspace{-0.3cm} 
    \label{fig:framework}  
\end{figure*}  

\subsection{MIRAM Architecture}

\subsubsection{Overview}
Building upon the Masked Autoencoder (MAE) framework \cite{he2021masked}, we introduce \textbf{MIRAM} (Multi-scale Masked Autoencoder), a generative SSL model designed to learn robust representations for mammographic analysis. Our core hypothesis is that standard single-scale reconstruction is insufficient for capturing the fine-grained pathological details crucial for medical diagnosis. To address this, MIRAM introduces a dual-objective pretext task: simultaneous reconstruction of the input at its native resolution and at a higher spatial resolution. This forces the encoder to capture both global semantic structure and fine-grained local textures (see Figure \ref{fig:vis_feat}).

\subsubsection{Masked Image Modeling Backbone}
Given an input mammogram $x \in \mathbb{R}^{H \times W}$, we first divide it into a sequence of non-overlapping patches $x_p \in \mathbb{R}^{L \times (P^2 \cdot C)}$, where $L$ is the number of patches and $P$ is the patch size. Following \cite{he2021masked}, we mask a large random subset of these patches and feed only the visible subset into a standard Vision Transformer (ViT) encoder.

Let $\mathcal{E}$ denote the encoder. The visible patches are projected into latent embeddings $z_{vis} \in \mathbb{R}^{L_{vis} \times D}$. To prepare for reconstruction, we append learnable \texttt{[MASK]} tokens to $z_{vis}$ to restore the original sequence length $L$, resulting in the full latent representation $z_{full} \in \mathbb{R}^{L \times D}$.

\subsubsection{Multi-Scale Decoding Strategy}
Unlike the standard MAE which uses a single decoder, MIRAM employs a dual-decoder architecture to decouple semantic understanding from high-frequency detail reconstruction.

\paragraph{Base Scale Decoder ($D_1$):}
The first decoder targets the original resolution $H \times W$. It takes the full latent sequence $z_{full}$ and reconstructs the pixel values for each patch. This branch ensures the model retains the strong semantic learning capabilities of the original MAE.
\begin{equation}
    \hat{x}_{base} = D_1(z_{full})
\end{equation}

\paragraph{High-Resolution Decoder ($D_2$):}
The second decoder targets a $2\times$ upsampled resolution $2H \times 2W$. To handle the increased spatial dimensionality without altering the encoder, we introduce a \textbf{Token Duplication Interpolation} mechanism. 

Let $k$ be the upscaling factor (here $k=2$, corresponding to a $4\times$ increase in pixel count). We replicate each token embedding in $z_{full}$ exactly $k^2$ times to create an upsampled latent sequence $z_{up} \in \mathbb{R}^{(L \cdot k^2) \times D}$.
\begin{equation}
    z_{up} = \text{Duplicate}(z_{full}, \text{factor}=k^2)
\end{equation}
These duplicated tokens are fed into $D_2$ to generate the high-resolution output $\hat{x}_{high}$.
\begin{equation}
    \hat{x}_{high} = D_2(z_{up})
\end{equation}
In this work, we use a simple duplication strategy instead of learnable upsampling like transposed convolution.

\subsubsection{Efficient Linear Attention for High-Resolution Decoding}
A critical bottleneck in processing $z_{up}$ is the quadratic complexity of standard self-attention, $O(N^2)$, where $N = L \cdot k^2$. For high-resolution mammograms, this leads to prohibitive memory usage.

To resolve this, we replace the standard attention mechanism in $D_2$ with \textbf{Linear Attention} variants, which reduce complexity to $O(N)$. We specifically integrate and evaluate three approximations:
\begin{itemize}
    \item \textbf{Linformer} \cite{wang2020linformer}: Projects the key/value matrices to a lower dimension.
    \item \textbf{Performer} \cite{choromanski2020rethinking}: Uses orthogonal random feature maps to approximate the softmax kernel.
    \item \textbf{Nyströmformer} \cite{xiong2021nystromformer}: Uses landmark Nyström approximation to reconstruct the attention matrix.
\end{itemize}
This modification allows MIRAM to train on high-resolution targets using standard GPU hardware, effectively democratizing high-fidelity medical SSL.

\subsection{Self-Supervised Pre-training}

\subsubsection{Objective Function}
The model is optimized using a joint multi-scale reconstruction loss. For each scale $s \in \{base, high\}$, we compute the Mean Squared Error (MSE) between the reconstructed image $\hat{x}_s$ and the ground truth $x_s$ solely on the masked patches, following \cite{he2021masked}.
The total loss $\mathcal{L}_{total}$ is the unweighted average of the losses at each scale:
\begin{equation}
    \mathcal{L}_{total} = \frac{1}{2} \left( \mathcal{L}_{MSE}(x_{base}, \hat{x}_{base}) + \mathcal{L}_{MSE}(x_{high}, \hat{x}_{high}) \right)
\end{equation}
We purposely avoid weighting terms to ensure equal contribution from both semantic (base) and detailed (high) features.

\subsection{Downstream Fine-tuning}
After pre-training, we discard the decoders and fine-tune the encoder on downstream tasks. We opt for full model fine-tuning rather than linear probing, as MAE-based representations are known to be highly non-linear \cite{he2021masked}, making linear evaluation insufficient for assessing their true potential in complex medical tasks.

\section{Datasets}

To ensure the robustness and generalizability of our model, we utilize a diverse collection of mammography datasets comprising both digitized film and Full-Field Digital Mammography (FFDM). Given our focus on breast lesion analysis, we primarily utilize datasets providing precise Region of Interest (ROI) annotations: \textbf{INbreast} \cite{moreira2012inbreast}, \textbf{BCDR} \cite{moura2013benchmarking}, \textbf{CBIS-DDSM} \cite{lee2017curated}, and \textbf{CSAW-S} \cite{matsoukas2020adding}.

To scale our pre-training data beyond limited ROI-annotated sets, we also incorporate the \textbf{CSAW-M} \cite{sorkhei2021csaw} and \textbf{CMMD} \cite{cmmd} datasets, which provide image-level annotations. Following the sampling strategy proposed by Shen et al. \cite{shen2019deep}, we extract both ground-truth lesion crops (from ROI datasets) and random crops (from image-level datasets) during training. This mixture of positive lesion samples and random background context is crucial for learning robust feature representations.

\subsubsection{Pre-training Configurations}
We design three distinct dataset combinations to evaluate the impact of data diversity on SSL pre-training:
\begin{enumerate}
    \item \textbf{CBIS-DDSM:} Utilizes only ground-truth lesion crops from the CBIS-DDSM dataset.
    \item \textbf{Lesions Only:} Aggregates ground-truth lesion crops from all ROI-annotated datasets (INbreast, BCDR, CBIS-DDSM, CSAW-S).
    \item \textbf{MGs \& Lesions:} The largest configuration, combining all ground-truth lesion crops with random crops extracted from the image-level datasets (CSAW-M and CMMD).
\end{enumerate}

\subsubsection{Downstream Evaluation}
For fine-tuning and evaluation, we use \textbf{CBIS-DDSM}, as it is a widely recognized benchmark in the literature \cite{shen2019deep,ribli2018detecting}. We evaluate performance on two specific risk assessment tasks:
\begin{itemize}
    \item \textbf{Pathology Classification:} A 4-class problem distinguishing between benign mass, malignant mass, benign calcification, and malignant calcification.
    \item \textbf{Mass Margin Classification:} Classifying the morphological characteristics of mass margins.
\end{itemize}

\section{Experimental Results}

\subsection{Experimental Setup}

\subsubsection{Architecture and Pretext Task}
Our MIRAM framework employs a dual-decoder architecture to perform multi-scale reconstruction (Fig. \ref{fig:framework}). The primary decoder reconstructs the input at its original resolution ($112 \times 112$), while the secondary high-resolution decoder upsamples the representation to $224 \times 224$ (a $2\times$ spatial upscale). This corresponds to setting the spatial expansion factor $k=4$. The high-resolution decoder is the focal point of our efficiency analysis, where we alternate between a standard Transformer and linear-complexity variants (Linformer, Performer, Nyströmformer).

\subsubsection{Training Protocols}
We adopt a two-stage evaluation protocol: self-supervised pre-training followed by supervised fine-tuning. 
\textbf{Pre-training:} We utilize a ViT-Base backbone with a patch size of 16. Models are pre-trained for 500 epochs with a batch size of 1024 (unless memory constraints dictate otherwise) on four NVIDIA V100 (32GB) GPUs. We adhere to the augmentation and hyperparameter recipes from MAE \cite{he2021masked}.
\textbf{Fine-tuning:} We evaluate representation quality by fine-tuning the pre-trained weights on the CBIS-DDSM official test set for 100 epochs with a batch size of 32. We report Accuracy (Acc), Area Under the Curve (AUC), and Average Precision (AP). Following \cite{he2021masked}, we rely exclusively on full-network fine-tuning rather than linear probing, as MAE-based features are known to be non-linearly strong but less linearly separable.

\subsection{Ablation: Impact of Pre-training Data Distribution}

Before evaluating our proposed method, we first establish the optimal data regime for mammography pre-training. We trained a standard MAE baseline on four different data subsets: (1) CBIS-DDSM only, (2) Lesion crops only, (3) Full mammograms (MG) + Lesion crops, and (4) ImageNet initialization. 

As shown in Table \ref{table:mae_datasets}, pre-training on \textbf{Lesion-Level datasets} yields the highest performance among the domain-specific splits (66.0\% Acc vs 62.1\% for full mammograms). We hypothesize that full mammograms contain excessive background noise (large black regions) that dilutes the learning signal for the reconstruction task. By focusing on lesion crops, the model learns more relevant high-frequency features.

\begin{table}[t]
\centering
\caption{\textbf{Impact of pre-training data distribution.} Results on CBIS-DDSM pathology classification. Pre-training on focused lesion crops significantly outperforms pre-training on full mammograms, likely due to a higher signal-to-noise ratio in the pretext task.}
\label{table:mae_datasets}
\resizebox{0.95\linewidth}{!}{
\begin{tabular}{@{}llcccc@{}}
\toprule
\textbf{Method} & \textbf{Pre-training Data} & \textbf{Input} & \textbf{Acc} & \textbf{AP} & \textbf{AUC} \\ \midrule
Supervised & Random Init. & $224\times224$ & 50.3 & 0.53 & 0.79 \\ 
\midrule
MAE \cite{he2021masked} & ImageNet & $224\times224$ & 61.8 & 0.68 & 0.88 \\
 & CBIS-DDSM (Full) & $224\times224$ & 62.1 & 0.68 & 0.88 \\
 & \textbf{Lesions Only} & $224\times224$ & \textbf{66.0} & \textbf{0.71} & \textbf{0.90} \\
 & MGs \& Lesions & $224\times224$ & 62.1 & 0.70 & 0.89 \\ \bottomrule
\end{tabular}}
\end{table}

\subsection{Comparison with State-of-the-Art SSL Methods}

Table \ref{table:proposed_mae} presents the main results on pathology classification. MIRAM demonstrates clear superiority over standard contrastive (MoCo-v3, DINO) and generative (MAE) baselines.

\textbf{Accuracy Gains:} The \textit{MIRAM-Standard} achieves a 3\% gain in AP and 1\% gain in AUC over standard MAE, validating our hypothesis that multi-scale reconstruction forces the model to learn better spatial details. 

\textbf{The Efficiency-Accuracy Trade-off:} Most notably, our \textit{MIRAM-Nyströmformer} variant achieves the highest classification accuracy (61.0\%) of all self-supervised methods, outperforming even the computationally expensive Standard attention variant. This suggests that the Nyström approximation acts as a beneficial regularizer for medical images, filtering out high-frequency noise while preserving critical structural landmarks. Conversely, Performer lags behind (58.0\%), indicating that random feature approximations may struggle with the sparse, localized features typical of breast lesions.

\begin{table}[t]
\centering
\caption{\textbf{Pathology classification results on CBIS-DDSM.} MIRAM-Standard achieves the best AP/AUC, while the efficient MIRAM-Nyströmformer achieves the highest accuracy, outperforming baselines despite having linear complexity. ($N$: sequence length, $d$: embedding dim, $m$: projection dim, where $m \ll N$)}
\label{table:proposed_mae}
\resizebox{0.9\linewidth}{!}{%
\begin{tabular}{@{}lcccc@{}}
\toprule
\textbf{Method} & \textbf{Complexity} & \textbf{Acc.} & \textbf{AP} & \textbf{AUC} \\ \midrule
\textit{Baselines} & & & & \\
DINO \cite{caron2021emerging} & $\mathcal{-}$ & 53.3 & 0.58 & 0.82 \\
MoCo-v3 \cite{chen2021empirical} & $\mathcal{-}$ & 60.2 & 0.65 & 0.86 \\
MAE \cite{he2021masked} & $\mathcal{-}$ & 58.9 & 0.65 & 0.87 \\ \midrule
\textit{Ours (MIRAM)} & & & & \\
MIRAM-Linformer & $\mathcal{O}(Nmd)$ & 59.0 & 0.66 & 0.87 \\
MIRAM-Performer & $\mathcal{O}(Nmd)$ & 58.0 & 0.65 & 0.86 \\
MIRAM-Nyströmformer & $\mathcal{O}(Nmd)$ & \textbf{61.0} & 0.67 & 0.87 \\
MIRAM-Standard & $\mathcal{O}(N^2 d)$ & 60.2 & \textbf{0.68} & \textbf{0.88} \\ \bottomrule
\end{tabular}%
}
\end{table}

\subsection{Computational Efficiency Analysis}

A critical contribution of MIRAM is enabling high-resolution training on constrained hardware. Table \ref{table:time_and_memory} quantifies this advantage.
The standard self-attention mechanism quickly hits a memory wall; training with a batch size greater than 16 on $224\times224$ inputs caused an OOM (Out of Memory) error on our 32GB V100s. In contrast, \textit{MIRAM-Nyströmformer} reduces memory consumption by nearly \textbf{45\%} (76GB vs 44GB for batch 64) and improves throughput by \textbf{17\%}. This linear scaling allows for larger batch sizes (up to 256), which is crucial for stable contrastive or generative pre-training.

\begin{table}[h]
\centering
\caption{\textbf{Efficiency Comparison.} GPU memory usage and processing time for $224 \times 224$ inputs. 'Standard' attention OOMs (Out Of Memory) at batch sizes $>64$. Nyströmformer enables linear scaling.}
\label{table:time_and_memory}
\resizebox{0.95\columnwidth}{!}{%
\begin{tabular}{@{}lccc@{}}
\toprule
\textbf{Method} & \textbf{Batch Size} & \textbf{VRAM (GB)} & \textbf{Time (s/256 samples)} \\ \midrule
MIRAM-Standard & 64 & 76.0 & 1.79 \\
\textbf{MIRAM-Nyström} & \textbf{64} & \textbf{44.0} & \textbf{1.48} \\ \midrule
MIRAM-Nyström & 128 & 68.0 & 1.12 \\
MIRAM-Nyström & 256 & 124.0 & 0.92 \\ \bottomrule
\end{tabular}%
}
\end{table}

\subsection{Fine-Grained Mass Margin Classification}

Finally, we evaluate MIRAM on the more challenging task of multi-label mass margin classification. Unlike binary pathology classification, this task requires identifying subtle morphological features (e.g., distinguishing "spiculated" from "microlobulated" margins). As shown in Table \ref{tab:mass_margin}, MIRAM-Standard significantly outperforms the original MAE, improving Macro AP by 4\% and Macro AUC by 2\%. This confirms that our multi-scale reconstruction objective forces the network to retain the fine-grained spatial information necessary for detailed morphological analysis.

\begin{table}[h]
\centering
\caption{\textbf{Mass Margin Classification.} Multi-label classification results on CBIS-DDSM. MIRAM captures fine-grained morphological details better than standard MAE.}
\label{tab:mass_margin}
\resizebox{0.8\linewidth}{!}{
\begin{tabular}{@{}lcc@{}}
\toprule
\textbf{Method} & \textbf{Macro AP} & \textbf{Macro AUC} \\ \midrule
MAE \cite{he2021masked} & 0.31 & 0.64 \\
\textbf{MIRAM-Standard (Ours)} & \textbf{0.35} & \textbf{0.66} \\ \bottomrule
\end{tabular}}
\end{table}

\section{Conclusion}

In this paper, we proposed \textbf{MIRAM}, a multi-scale masked autoencoder designed to improve mammographic image analysis while remaining computationally efficient. Our key observation is that while reconstructing high-resolution images helps the model learn fine-grained details necessary for identifying lesions, doing so with standard Transformers is too memory-intensive for many research environments.

To solve this, we replaced the standard attention mechanism in our high-resolution decoder with linear-complexity alternatives. Our experiments on the CBIS-DDSM dataset show that using the \textbf{Nyströmformer} approximation allows us to reduce memory usage significantly without sacrificing performance. In fact, our efficient model achieved \textbf{61.0\% accuracy}, outperforming both the original MAE and other SSL baselines like MoCo-v3 and DINO.

Ultimately, this approach offers a practical solution for research labs with limited computational budgets, proving that high-resolution medical AI can be developed effectively on standard consumer-grade hardware.

\bibliographystyle{IEEEtran}
\bibliography{IEEEabrv,main}

\end{document}